\documentclass[conference]{IEEEtran}
\IEEEoverridecommandlockouts

\usepackage{cite}
\usepackage{amsmath,amssymb,amsfonts}
\usepackage{algorithmic}
\usepackage{graphicx}
\usepackage{textcomp}
\usepackage{xcolor}
\def\BibTeX{{\rm B\kern-.05em{\sc i\kern-.025em b}\kern-.08em
    T\kern-.1667em\lower.7ex\hbox{E}\kern-.125emX}}
\usepackage{booktabs}
\usepackage{multirow}
\usepackage{siunitx}
\usepackage{graphicx}
\usepackage{tikz}
\usetikzlibrary{calc}
\usepackage{verbatim}
\begin{document}

\title{PCB-MC: Missing Component Analysis in Printed Circuit Boards}
\author{\IEEEauthorblockN{1\textsuperscript{st} Betsy Villa}
\IEEEauthorblockA{\textit{EEMCS-CS-DMB} \\
\textit{University of Twente}\\
Enschede, The Netherlands\\
b.j.villabrochero@utwente.nl}
\and
\IEEEauthorblockN{2\textsuperscript{nd} Ian Gibson}
\IEEEauthorblockA{\textit{ET-DPM-AMSPES} \\
\textit{University of Twente}\\
Enschede, The Netherlands\\
i.gibson@utwente.nl}
\and
\IEEEauthorblockN{3\textsuperscript{rd} Estefania Talavera}
\IEEEauthorblockA{\textit{EEMCS-CS-DMB} \\
\textit{University of Twente}\\
Enschede, The Netherlands\\
e.talaveramartinez@utwente.nl }
}
\maketitle

\begin{abstract}
Detecting missing components on printed circuit boards (PCBs) differs fundamentally from conventional object detection, as the model must localize components that are not present. We introduce PCB-MC, a curated dataset for missing component detection with footprint level annotations built on top of the RF100 dataset. The dataset contains 197 distinct board types, each corresponding to a unique PCB design, with multiple augmented samples per type. We also provide benchmark results on PCB-MC by evaluating a diverse set of supervised and unsupervised methods. To ensure fair evaluation, we propose board type aware cross validation splits that prevent layout leakage between training and test sets. Supervised models showcase high false negative rates on unseen board designs, and unsupervised anomaly detection methods fail entirely due to the lack of spatial alignment with a board specific reference. These results confirm that missing component detection on diverse PCB layouts remains an open challenge. We release PCB-MC and all training protocols to support reproducible research on structural absence detection in industrial inspection.
\end{abstract}

\begin{IEEEkeywords}
Missing Component Detection, Industrial Visual Inspection, Object Detection, and Dataset.
\end{IEEEkeywords}

\section{Introduction}
\label{sec:intro}

Automated visual inspection is critical in Printed Circuit Board (PCB) manufacturing, where undetected defects can lead to device failure and costly rework \cite{Ling2023}. While substantial progress has been made in detecting visible components and surface defects, missing component detection remains comparatively underexplored. 

\begin{figure}[!h]
\centering
\includegraphics[width=0.7\linewidth]{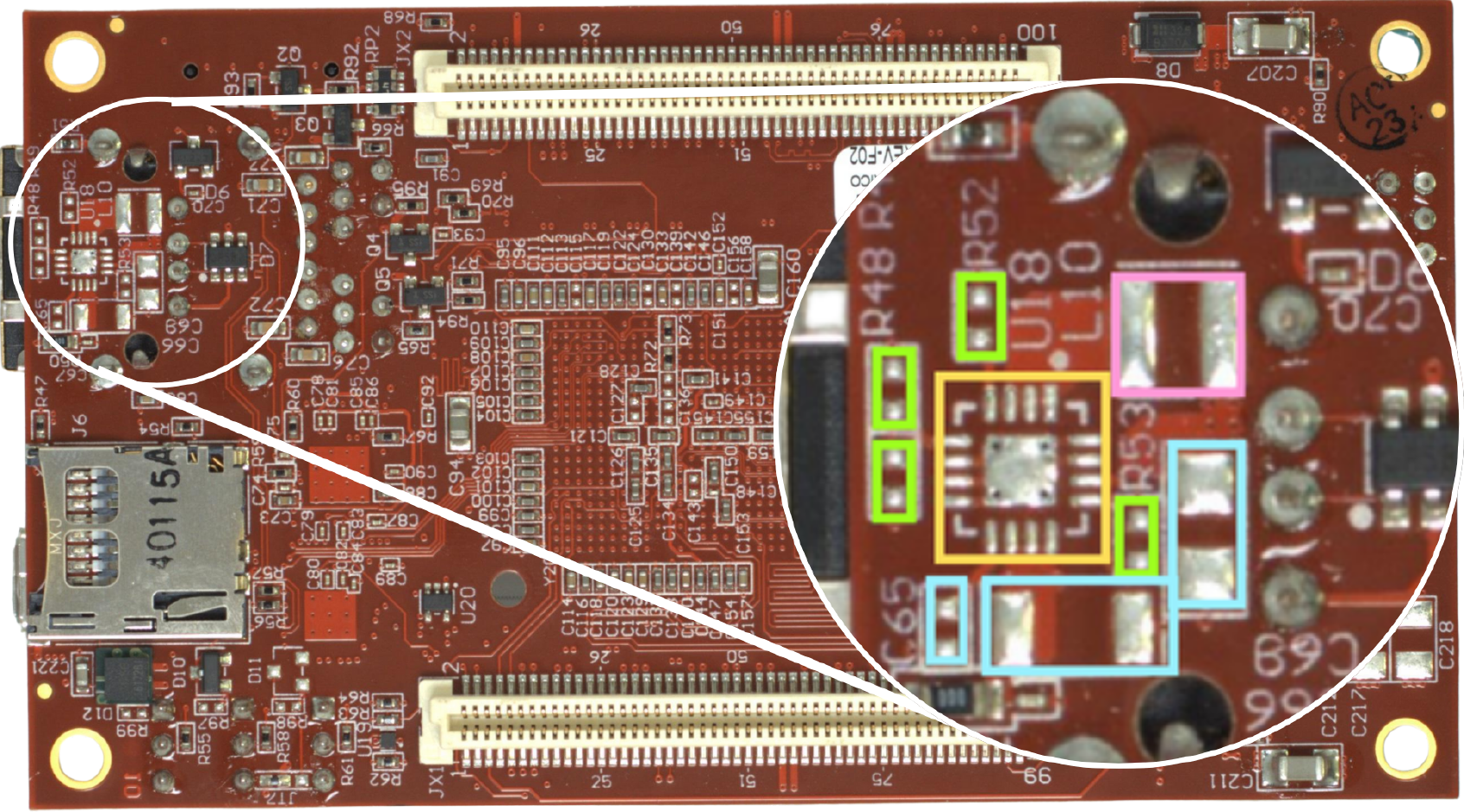}
\includegraphics[width=0.7\linewidth]{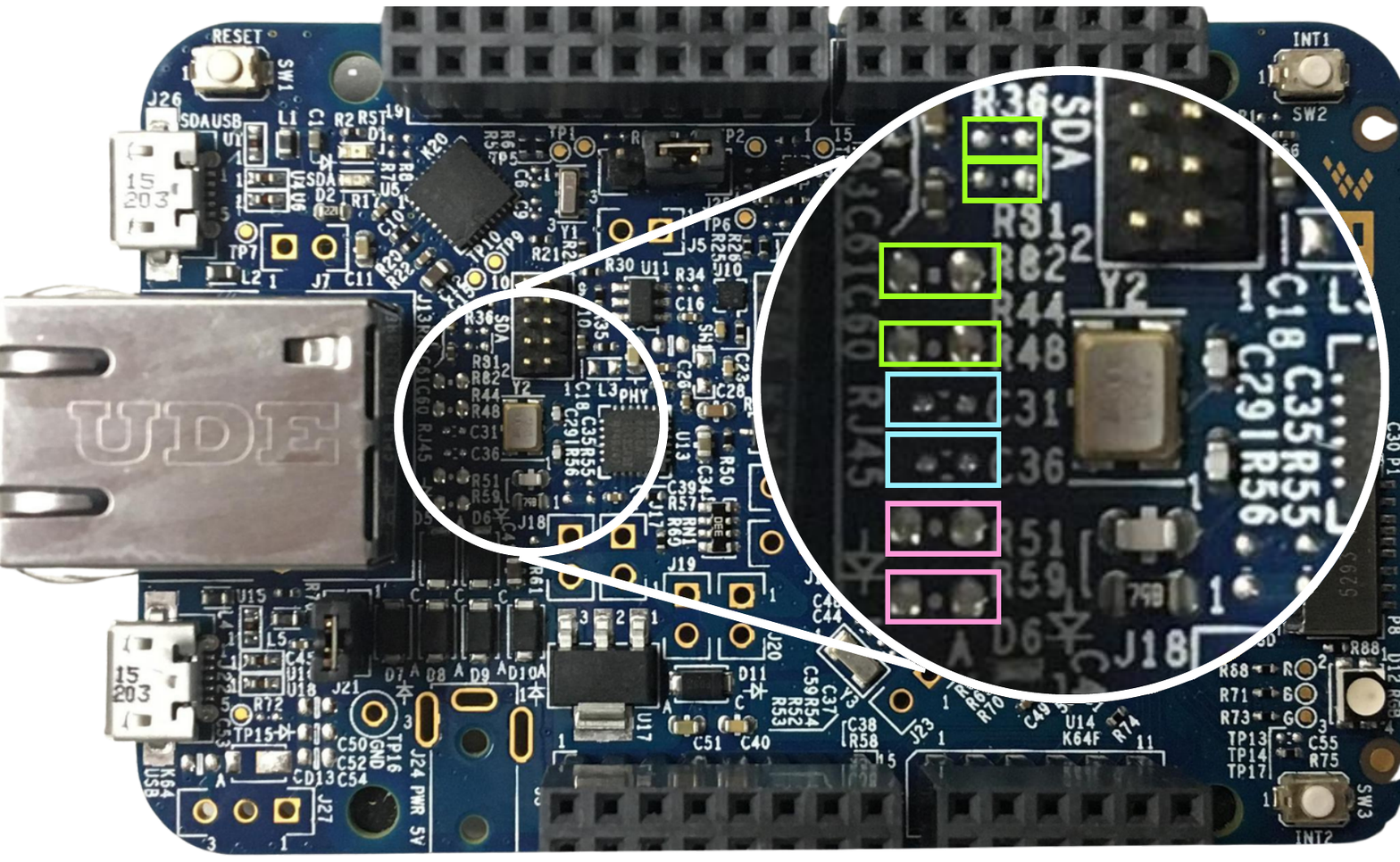}
\caption{Illustration of missing component annotations in PCB-MC. Color coded bounding boxes denote some missing components: capacitors (blue), resistors (green), ICs (yellow), and LEDs (pink). The zoomed region highlights subtle footprint cues such as exposed solder pads and silkscreen outlines.}
\label{fig:missingpartfig1}
\end{figure}

Unlike conventional object detection, missing component detection in PCBs requires inferring the expected footprint of an absent component based on subtle structural cues such as exposed solder pads and silkscreen outlines (Figure~\ref{fig:missingpartfig1}). This distinction is not captured by existing PCB benchmarks, which focus on component recognition or visible defect detection and lack annotations for missing components at their expected locations, making them unsuitable for evaluating absence based detection.

In this work, we introduce \textbf{PCB-MC}, a dataset for missing component detection built on RF100~\cite{RF100}. PCB-MC provides footprint level annotations for absent components across 615 images and 31 categories, comprising over \textbf{9,400 missing instances}. The dataset spans 197 distinct types of boards, each corresponding to a unique PCB layout, with an average of 3.12$\pm$1.02  samples per type. Missing components are labeled at their expected locations, enabling direct supervision of structural absence.

We benchmark modern detection models \cite{yolov8_ultralytics,yolo11_ultralytics,yolo26_ultralytics,zhao2024detrsbeatyolosrealtime,peng2024dfineredefineregressiontask} alongside unsupervised anomaly detection methods \cite{roth2022patchcore,defard2021padim,zavrtanik2021draemdiscriminativelytrained,deng2022reverse}.  In addition, we establish a controlled evaluation framework with board type aware splits to prevent layout memorization and high resolution inputs ($1024 \times 1024$) to preserve fine grained structural cues, ensuring fair comparisons under realistic industrial conditions. 

Beyond benchmarking, we conduct a controlled analysis to better understand the challenges of missing component detection, particularly examining the roles of candidate region generation and classification under weak visual evidence.

The PCB-MC annotations and the evaluation protocol will be made publicly available\footnote{Available upon acceptance} to support reproducible research on structural absence detection in industrial inspection.

\section{Related Work}
\label{sec:related}
\subsection{Computer Vision in Industrial Inspection}
Most studies in this field have focused on anomaly detection, where models learn normal patterns and detect deviations \cite{liu2024deep,bergmann2019mvtec}. Methods such as \cite{roth2022patchcore}, \cite{defard2021padim}, \cite{zavrtanik2021draemdiscriminativelytrained}, and \cite{deng2022reverse} perform well on industrial anomaly benchmarks, but are primarily designed for visible defects. However, these approaches are less effective for complex production lines with small components or limited training data, where defining 'normal' behavior is significantly more challenging. An example is PCB assembly. As electronics become more common and advanced, PCBs are getting more complex, with smaller parts and denser layouts that are much harder for standard models to inspect.

\subsection{PCB Inspection and Defect Detection}
In practical applications, production environments exhibit considerable variability. High volume production lines typically employ automated assembly and inspection processes. However, numerous low and medium volume productions such as prototyping, specialized industrial boards, repair, and small batch manufacturing continue to rely on human assembly or reworking. In these contexts, the implementation of an automated system for detecting defects, particularly those involving missing components, is crucial. Such systems are essential for preventing latent failures, minimizing the need for manual reinspection, and enhancing consistency among operators \cite{Suksukont2025}.

Early PCB inspection relied on classical techniques such as template matching \cite{Demir1994} and image differencing \cite{Chavan2016QualityCO}, which are highly sensitive to misalignment and illumination changes.

Recent advances in deep learning have significantly improved robustness and scalability in PCB inspection. Convolutional neural network (CNN) detectors, particularly YOLO \cite{yolov8_ultralytics,yolo11_ultralytics,yolo26_ultralytics} style architectures, are widely adopted for real time component localization and defect detection \cite{Chhetri2023,xia_global_2023}.Transformer base detectors such as RT-DETR \cite{Tang2025LRTDETRAE} further enhance global context modeling, which can be beneficial for complex board layouts \cite{zhao2024detrsbeatyolosrealtime}. 

Two stage detectors, including Faster R-CNN \cite{FasterRCNN} and Mask R-CNN \cite{MaskRCNN}, have also been applied to PCB defect detection, often achieving strong localization performance at the cost of higher computational overhead \cite{calabrese_application_2025}. Prior PCB specific work has focused on enhancing small object detection \cite{Hu2020PCBFasterRCNN} and improving detection flexibility via architectural modifications \cite{Luo2021DecoupledTwoStage}. More recent approaches incorporate contextual reasoning and attention mechanisms to further boost performance \cite{Kiobya2024ACASEM}.

Despite these advances, most existing work focuses on detecting visible components or surface level defects (e.g., scratches, solder bridges, missing holes). Multi class missing component detection remains significantly more challenging than single defect scenarios, highlighting the need for dedicated datasets and systematic evaluation protocols.

\subsection{PCB Datasets}

Previous works on PCB inspection have been defined by the provided labels in the publicly available datasets. RF100 \cite{RF100}, FPIC \cite{Makwana2023}, FICS-PCB \cite{Lu2020}, PCB-Metal \cite{Mahalingam2015}, and PCB analysis \cite{Pramerdorfer2015} are primarily designed for component recognition tasks. Defect oriented datasets such as DeepPCB \cite{Tang2018}, PCB Defect \cite{Li2019}, and DsPCBSD \cite{LV2024} focus on visible surface anomalies or template comparison approaches \cite{Savu2025}. While these datasets support defect detection, they do not explicitly model missing components at their expected locations. PCB-MC provides new labels for the task of missing components detection, which provides a new benchmark for the area.

\section{PCB-MC: A Dataset for Missing Component Detection}
\label{sec:dataset}
PCB-MC is derived from the “Printed Circuit Board” dataset from Roboflow Universe (RF100) \cite{RF100}, which contains 615 images with diverse layouts and component types. \textbf{PCB-MC} extends RF100 with annotations for missing components at their expected footprint locations. Moreover, all annotations were manually reviewed. We corrected inaccurate bounding box localizations, standardized label names, merged visually equivalent subclasses, and removed ambiguous categories to improve annotation consistency.

After our refinement and annotation protocol, our newly introduced PCB-MC dataset contains \textbf{615 images} annotated across \textbf{31 classes}. These 31 classes include 23 of the present components, 8 of which are also identified as absent. Table~\ref{tab:class_stats} presents the instance distribution across present and missing categories. The dataset exhibits substantial class imbalance, with passive components such as resistors and capacitors dominating both present and missing annotations. This imbalance reflects realistic industrial assembly conditions, where small passive components are both frequent and prone to omission.

In addition to instance counts, missing components are predominantly small in spatial extent relative to the full board image resolution. This motivates high resolution evaluation settings to preserve fine grained footprint cues.

\begin{table}[t]
\caption{Component class distribution in the PCB-MC dataset.}
\label{tab:class_stats}
\centering
\scriptsize
\begin{tabular}{l
                S[table-format=2.0]
                S[table-format=6.0]
                S[table-format=5.0]}
\toprule
Class & {\#Classes} & {\#Present} & {\#Missing} \\
\midrule
Button        & 1 & 235    & 0 \\
Capacitor     & 2 & 46377  & 3177 \\
Clock         & 1 & 121    & 0 \\
Connector     & 1 & 4011   & 0 \\
Diode         & 2 & 212    & 146 \\
Display       & 1 & 17     & 0 \\
Electrolytic Cap. & 1 & 697 & 0 \\
EM            & 1 & 138    & 0 \\
Ferrite Bead  & 2 & 316    & 42 \\
Fuse          & 1 & 23     & 0 \\
Heatsink      & 1 & 14     & 0 \\
IC            & 2 & 6845   & 169 \\
Inductor      & 2 & 192    & 72 \\
Jumper        & 1 & 291    & 0 \\
LED           & 2 & 677    & 19 \\
Pads          & 1 & 327    & 0 \\
Pins          & 1 & 1028   & 0 \\
Potentiometer & 1 & 25     & 0 \\
Resistor      & 2 & 50355  & 4450 \\
Switch        & 1 & 165    & 0 \\
Test Point    & 1 & 1108   & 0 \\
Transistor    & 1 & 3904   & 0 \\
Unknown       & 1 & 0      & 1360 \\
Zener Diode   & 1 & 13     & 0 \\

\midrule
\textbf{Total} & 31 & 117091 & 9435 \\
\bottomrule
\end{tabular}
\end{table}

\textit{Annotation Protocol for Missing Components:}
Missing components are annotated at expected footprint locations using cues such as solder pads, silkscreen outlines, and layout regularities. All annotations are cross validated by multiple annotators.
In total, PCB-MC contains \textbf{117,091 present component instances} and \textbf{9,435 missing component instances}. Missing components are annotated at their expected footprint locations, enabling supervised learning for absence localization.

\section{Benchmark Protocol on PCB-MC}
\label{sec:Method}
We described the tasks that we propose on top of our newly introduced PCB-MC dataset, the implemented methods for missing object detection, and the designed evaluation protocol.

\subsection{Benchmark Tasks}

PCB-MC supports multiple evaluation settings derived from the unified annotation set:

\begin{itemize}
    \item \textbf{Task A (All Classes):} Joint detection of present and missing components across all 31 categories (615 images).
    \item \textbf{Task M (Missing Only):} Detection restricted to the 8 missing component classes (293 images).
    \item \textbf{Task C (Components Only):} Detection restricted to the 23 present component classes (615 images).
\end{itemize}

These tasks enable controlled comparison between standard component detection and structural absence detection under consistent annotation conditions. Table~\ref{tab:subset} summarizes the dataset composition under each benchmark configuration.
\begin{table}[t]
\caption{PCB-MC dataset splits and their statistics.}
\label{tab:subset}
\centering
\scriptsize
\begin{tabular}{l l
                S[table-format=2.0]
                S[table-format=3.0]}
\toprule
Subset & Class Type & {\#Cls} & {\#Imgs} \\
\midrule
PCB-MC-A & Components + Missing & 31 & 615 \\
PCB-MC-M & Missing parts only   & 8  & 293 \\
PCB-MC-C & Components only      & 23 & 615 \\
\bottomrule
\end{tabular}
\end{table}

\subsection{Missing Component Detection Methods}
\paragraph{Supervised  Object Detection Models}
We evaluate five object detection architectures covering complementary paradigms, including convolutional detectors YOLOv8/11/26 \cite{yolov8_ultralytics,yolo11_ultralytics,yolo26_ultralytics} and transformer based models RT-DETR \cite{Tang2025LRTDETRAE} and D-FINE \cite{peng2024dfineredefineregressiontask}. We additionally evaluate tiled inference using SAHI~\cite{SAHI} to assess whether increased local detail improves detection. All models are trained under identical settings to ensure fair comparison. 

Furthermore, in this work, we design a two stage missing component detection pipeline that leverages object proposals by an object detector and a specialized classifier.
We rely on the YOLOv11 model trained on the PCB-MC-A training split for each fold, comprising all 31 classes across both present and missing component annotations for the first stage of object proposals.YOLOv11 is then run at a low confidence threshold ($\text{conf} = 0.05$) to generate candidate bounding boxes across the entire image without class filtering. At this threshold, the detector produces approximately 240 proposals per image, covering every region it considers potentially relevant.

In the second stage, we rely on the ResNet-18 backbone for classification ~\cite{DBLP:journals/corr/HeZRS15}. ResNet-18 is trained on proposals of present and missing components. We use focal loss ~\cite{lin2018focallossdenseobject}
($\alpha = 0.75$, $\gamma = 2.0$) to address the 2:1 present to missing class ratio. During training, each patch is expanded by a factor of $1.4\times$ to capture surrounding context such as pad geometry and silkscreen outlines. High confidence missing predictions are retained and filtered with non maximum suppression. 

\paragraph{Anomaly Detection Models}

Given the broad use of anomaly detection methods in industrial inspection, we assess their performance on the PCB-MC dataset. We evaluate the major anomaly detection families\cite{liu2024deep} PatchCore \cite{roth2022patchcore}, PaDiM~\cite{defard2021padim}, DRAEM \cite{zavrtanik2021draemdiscriminativelytrained}, and Reverse Distillation \cite{deng2022reverse}. All models are implemented using Anomalib \cite{akcay2022anomalib} and trained on images with boards including all present components, then tested on full images with boards with missing components.

Pixel level anomaly maps are converted to bounding boxes via thresholding (0.995 quantile to control false positives) \cite{bergmann2019mvtec}, connected component extraction with noise removal, and non maximum suppression (IoU\,=\,0.3) to merge overlapping detections.

\subsection{Evaluation}

Detection performance is evaluated using mAP@0.5 as the primary metric, following standard PCB detection benchmarks \cite{Luo2021DecoupledTwoStage, Mahalingam2015}, together with F1 score and False Negative Rate (FNR) to better assess missed detections. Given the industrial importance of avoiding missed defects, we emphasize the FNR for missing component classes.

PCB-MC contains 197 board types, where each board type corresponds to a unique PCB layout and component arrangement. The images are of high resolution and with small components. Therefore, the missing object detection methods are applied at the image patch level. 

To mitigate data leakage, we adopt a board type aware 5 fold cross validation protocol in which all image patches belonging to the same board type are assigned to the same fold. For each fold, 70\% of board types are used for training and 30\% for validation. Results are reported as mean $\pm$ standard deviation across folds without task specific hyperparameter tuning.

\subsection{Implementation Details}
All models are initialized with COCO pretrained weights ~\cite{lin2015microsoftcococommonobjects}, which provides diverse object level features that improve convergence and generalization. The pretrained weights are then fine tuned on PCB-MC for 300 epochs using AdamW~\cite{loshchilov2019decoupledweightdecayregularization} with an initial learning rate of $1 \times 10^{-4}$ and cosine scheduling.

To mitigate class imbalance and improve robustness to orientation and illumination variations, we apply data augmentation, including flipping, rotation, brightness/contrast adjustment, and mosaic augmentation. Inference uses a confidence threshold of 0.25 and class aware non maximum suppression.

Missing component footprints are small relative to the full board resolution, therefore all experiments are conducted with NVIDIA GPU acceleration at $1024 \times 1024$ resolution to preserve fine grained structural cues such as pads and silkscreen outlines, and configuration files are provided for reproducibility.

\begin{table}[t]
\caption{Detection performance across dataset subsets (mean $\pm$ std).
\textbf{Task M (missing components)} is the primary task of interest.}
\label{tab:results}
\centering
\scriptsize
\setlength{\tabcolsep}{4pt}
\renewcommand{\arraystretch}{1.2}

\begin{tabular}{l l c c c}
\toprule
\textbf{Method} & \textbf{Task} & \textbf{mAP} & \textbf{F1} & \textbf{FNR} \\
\midrule

YOLOv8~\cite{yolov8_ultralytics}
 & A & 0.29$\pm$0.05 & 0.37$\pm$0.03 & 0.69$\pm$0.04 \\
 & \textbf{M} & 0.09$\pm$0.13 & 0.15$\pm$0.02 & 0.90$\pm$0.02 \\
 & C & 0.37$\pm$0.07 & 0.46$\pm$0.05 & 0.51$\pm$0.07 \\
\midrule

YOLOv11~\cite{yolo11_ultralytics}
 & A & \textbf{0.30$\pm$0.05} & 0.39$\pm$0.06 & 0.67$\pm$0.05 \\
 & \textbf{M} & 0.09$\pm$0.03 & 0.15$\pm$0.04 & \textbf{0.87$\pm$0.03} \\
 & C & \textbf{0.42$\pm$0.09 }& \textbf{0.50$\pm$0.08} & 0.57$\pm$0.06 \\
\midrule

YOLO26~\cite{yolo26_ultralytics}
 & A & 0.29$\pm$0.03 & 0.36$\pm$0.04 & 0.69$\pm$0.02 \\
 & \textbf{M} & 0.08$\pm$0.04 & 0.15$\pm$0.05 & 0.87$\pm$0.05 \\
 & C & 0.37$\pm$0.07 & 0.47$\pm$0.03 & 0.61$\pm$0.07 \\
\midrule

YOLOv8 + SAHI~\cite{SAHI}
 & A & 0.27$\pm$0.05 & 0.45$\pm$0.05 & 0.51$\pm$0.08 \\
 & \textbf{M} & 0.06$\pm$0.03 & 0.13$\pm$0.07 & 0.92$\pm$0.05 \\
 & C & 0.34$\pm$0.13 & 0.46$\pm$0.05 & 0.53$\pm$0.05 \\

\midrule

YOLOv11 + SAHI~\cite{SAHI}
 & A & 0.29$\pm$0.05 & \textbf{0.47$\pm$0.05} & \textbf{0.50$\pm$0.09} \\
 & \textbf{M} & 0.08$\pm$0.02 & \textbf{0.16$\pm$0.05} & 0.90$\pm$0.04 \\
 & C & 0.36$\pm$0.10 & 0.48$\pm$0.04 & \textbf{0.51$\pm$0.04} \\
\midrule

YOLO26 + SAHI~\cite{SAHI}
 & A & 0.29$\pm$0.06 & 0.46$\pm$0.04 & 0.52$\pm$0.06 \\
 & \textbf{M} & 0.08 $\pm$0.03 & 0.13$\pm$0.05 & 0.92$\pm$0.03 \\
 & C & 0.38$\pm$0.07 & 0.46$\pm$0.08 & 0.58$\pm$0.08 \\
\midrule

RT-DETR~\cite{zhao2024detrsbeatyolosrealtime}
 & A & 0.28$\pm$0.06 & 0.37$\pm$0.05 & 0.69$\pm$0.05 \\
 & \textbf{M} & 0.04$\pm$0.03 & 0.11$\pm$0.05 & 0.92$\pm$0.04 \\
 & C & 0.38$\pm$0.09 & 0.46$\pm$0.08 & 0.58$\pm$0.08 \\
\midrule

D-FINE~\cite{peng2024dfineredefineregressiontask}
 & A & 0.24$\pm$0.03 & 0.25$\pm$0.03 & 0.75$\pm$0.03 \\
 & \textbf{M} & 0.03$\pm$0.02 & 0.04$\pm$0.02 & 0.93$\pm$0.02 \\
 & C & 0.31$\pm$0.08 & 0.32$\pm$0.07 & 0.67$\pm$0.07 \\
\midrule

Two stage & \textbf{M} & \textbf{0.15$\pm$0.05} & 0.13$\pm$0.04 & \textbf{0.75$\pm$0.07} \\
\midrule

PatchCore~\cite{roth2022patchcore} & M & 0.00$\pm$0.00 & 0.00$\pm$0.00 & 1.00$\pm$0.00 \\
PaDiM~\cite{defard2021padim} & M & 0.00$\pm$0.00 & 0.00$\pm$0.00 & 1.00$\pm$0.00 \\
DRAEM~\cite{zavrtanik2021draemdiscriminativelytrained} & M & 0.00$\pm$0.00 & 0.00$\pm$0.00 & 1.00$\pm$0.00 \\
Rev.\ Distill.~\cite{deng2022reverse} & M & 0.00$\pm$0.00 & 0.00$\pm$0.00 & 1.00$\pm$0.00 \\

\bottomrule
\end{tabular}
\end{table}

\begin{figure}[t]
\centering
\includegraphics[width=0.9\linewidth]{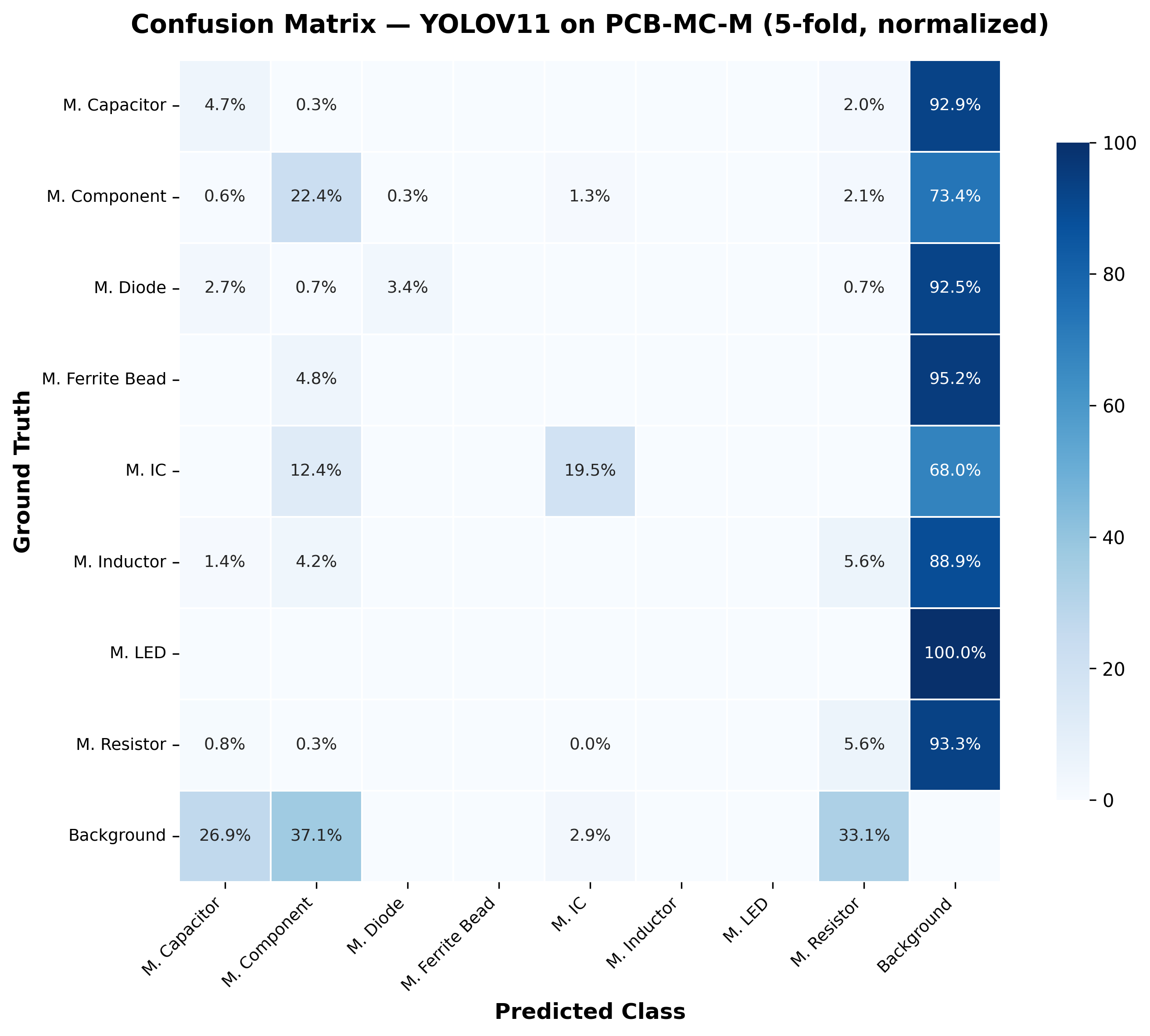}
\caption{
\textbf{Confusion matrix for YOLOv11 on PCB-MC-M (5-fold, normalized).}
Rows correspond to ground truth classes and columns to predicted classes.}
\label{fig:confusion_matrix}
\end{figure}

\section{Results and Discussion}
\label{sec:experiments}
\paragraph{Benchmark Task Performance}

The missing component task (M) constitutes the primary challenge addressed in this work and differs fundamentally from conventional object detection. Instead of recognizing visible appearance patterns, models must infer the absence of a component at an expected location using weak structural cues such as exposed pads or silkscreen outlines. As shown in Table~\ref{tab:results}, all one stage and transformer based detectors perform poorly on this task, with FNR values consistently above 0.87. YOLOv11 achieves the best one stage result (mAP 0.09, FNR 0.87), while RT-DETR and D-FINE perform substantially worse (FNR 0.92 and 0.93, respectively). Applying SAHI achieves the highest F1 on task (M), yet the FNR remains at 0.90, indicating that the limitation is not primarily related to scale or spatial resolution but to the inability of current detectors to represent structural absence.

In contrast, the proposed two stage approach achieves the best overall performance on task (M), with the highest mAP (0.15) and lowest FNR (0.75). Although its F1 (0.13) remains comparable to one stage detectors, the reduction in missed detections is critical in industrial inspection, where false negatives correspond to undetected assembly failures. These results suggest that explicitly separating proposal generation from classification provides a more suitable inductive bias for absence based detection.

The remaining tasks, All Classes (A) and Components only (C), follow conventional detection assumptions and achieve substantially stronger performance. In task (A), YOLOv11 reaches the highest mAP (0.30), while YOLOv11+SAHI obtains the best F1 (0.47) and lowest FNR (0.50), demonstrating the benefit of tiled inference for proposal coverage. Task (C) produces the strongest results overall, with YOLOv11 reaching mAP 0.42 and F1 0.50. Despite these improvements, FNR values remain relatively high across methods ($\approx$0.50--0.70), confirming that PCB inspection is challenging even under standard detection settings.
\begin{figure*}[t]
\centering
\scriptsize
\setlength{\tabcolsep}{2pt}
\renewcommand{\arraystretch}{1.1}

\newcommand{\cW}{0.15\textwidth}
\newcommand{\cH}{3cm}

\begin{tabular}{cccccc}
\scriptsize GT & \scriptsize Det & \scriptsize Prop &
\scriptsize GT & \scriptsize Det & \scriptsize Prop \\ \hline
\multicolumn{3}{c}{Sample 1} & \multicolumn{3}{c}{Sample 2} \\


\includegraphics[width=\cW,height=\cH,keepaspectratio]{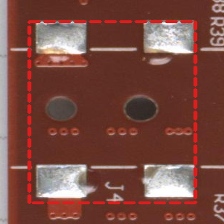} &
\includegraphics[width=\cW,height=\cH,keepaspectratio]{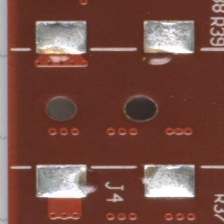} &
\includegraphics[width=\cW,height=\cH,keepaspectratio]{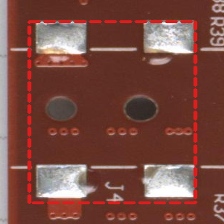} &
\includegraphics[width=\cW,height=\cH,keepaspectratio]{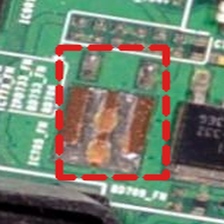} &
\includegraphics[width=\cW,height=\cH,keepaspectratio]{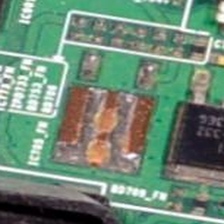} &
\includegraphics[width=\cW,height=\cH,keepaspectratio]{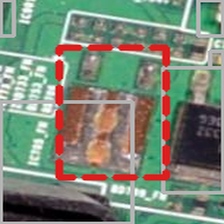} \\

\multicolumn{3}{c}{Sample 3} & \multicolumn{3}{c}{Sample 4} \\

\includegraphics[width=\cW,height=\cH,keepaspectratio]{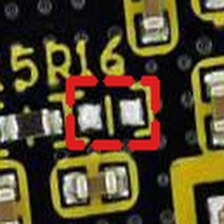} &
\includegraphics[width=\cW,height=\cH,keepaspectratio]{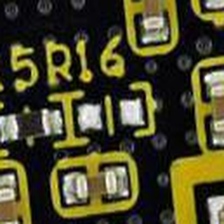} &
\includegraphics[width=\cW,height=\cH,keepaspectratio]{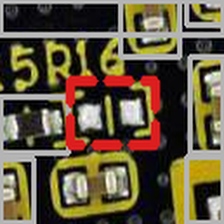} &
\includegraphics[width=\cW,height=\cH,keepaspectratio]{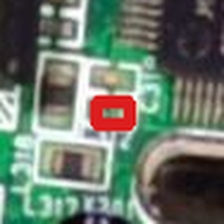} &
\includegraphics[width=\cW,height=\cH,keepaspectratio]{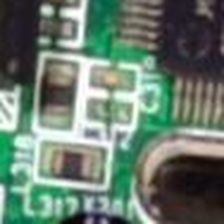} &
\includegraphics[width=\cW,height=\cH,keepaspectratio]{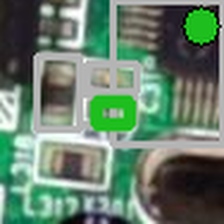} \\

\multicolumn{3}{c}{Sample 5} & \multicolumn{3}{c}{Sample 6} \\

\includegraphics[width=\cW,height=\cH,keepaspectratio]{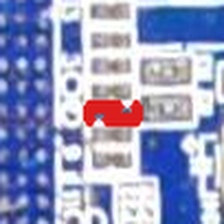} &
\includegraphics[width=\cW,height=\cH,keepaspectratio]{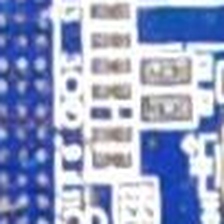} &
\includegraphics[width=\cW,height=\cH,keepaspectratio]{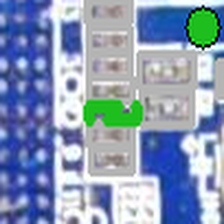} &
\includegraphics[width=\cW,height=\cH,keepaspectratio]{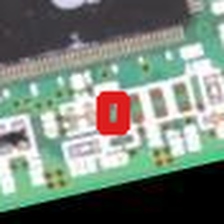} &
\includegraphics[width=\cW,height=\cH,keepaspectratio]{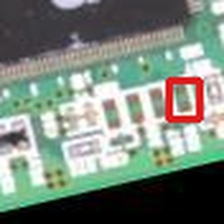} &
\includegraphics[width=\cW,height=\cH,keepaspectratio]{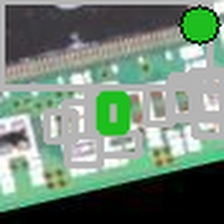} \\

\end{tabular}

\vspace{3pt}

\caption{%
\textbf{Component level diagnosis of detection failure on Task M.}
Each group shows a cropped missing component footprint at the inference scale.
\textbf{Columns:} Ground truth (red dashed box), detector output (YOLOv11, conf=0.25), and stage one proposals (conf=0.05).
\textbf{Samples 1-3:} missed due to no proposal coverage.
\textbf{Samples 4-6: } Correctly detected due to proposal coverage.
\textbf{Color legend:} red dashed = ground truth, gray = proposals, green = successful coverage.
}

\label{fig:component_crops}
\end{figure*}

\paragraph{Anomaly Detection vs. Object Detection}

Anomaly detection methods, including PatchCore, PaDiM, DRAEM, and Reverse Distillation, completely fail on the missing component task, producing zero detection performance and an FNR of 1.0. This behavior is expected, as anomaly detection methods assume consistent object structure across training and testing, whereas PCB-MC contains diverse board layouts without aligned references.

The failure can be attributed to a fundamental mismatch between the assumptions of anomaly detection and the structure of the PCB-MC task. These methods rely on learning the distribution of normal appearance and identifying deviations; however, missing components do not necessarily introduce strong local visual anomalies, especially in unaligned or layout variable settings. Instead, they require reasoning about expected object presence at specific spatial locations, which is not captured by appearance based anomaly modeling.

In comparison, object detection approaches implicitly capture spatial and semantic priors by means of bounding box supervision. However, they still struggle with missing components because training data typically lacks explicit negative examples of “expected but absent” objects. The two stage method partially overcomes this limitation by decoupling localization and verification, enabling more robust reasoning about absence.

\paragraph{Category Level Analysis}
Figure \ref{fig:confusion_matrix} presents the normalized confusion matrix for YOLOv11 on the task M. The dominant failure mode is not inter class confusion, but missed detections, where missing component instances are incorrectly assigned to the \textit{background} class (i.e., no detection produced for the corresponding footprint region). Across most categories, over 85\% of missing components are classified as background, with the error reaching 100\% for small footprints such as LEDs. 

In contrast, larger components with stronger structural cues, such as ICs, achieve comparatively higher detection rates, although the majority of instances are still missed (68.0\% predicted as background). The confusion matrix also shows minimal confusion between missing component categories, indicating that once a candidate region is generated, classification is comparatively less problematic than localization.

These results suggest that the primary limitation lies in identifying candidate regions corresponding to missing component footprints rather than discriminating between component types. Overall, the findings reinforce that missing component detection is fundamentally an absence localization problem requiring stronger spatial reasoning and structural priors beyond conventional appearance based detection.
\paragraph{Quantitative Evaluation}

The failure patterns in Figure~\ref{fig:component_crops} corroborate the results in Table~\ref{tab:results}. In Samples 1-3, one stage detectors fail to generate candidate regions over missing component footprints, confirming that these areas are not recognized as relevant objects. When proposal coverage is present (Samples 4-6), detection succeeds, which explains the improved FNR of the two stage method. Failures occur despite clear structural cues, and SAHI provides limited improvement, reinforcing that the bottleneck is semantic rather than resolution related.

\paragraph{Discussion} Taken together, these results demonstrate that PCB inspection is not a single homogeneous problem, but rather a spectrum ranging from pure appearance detection (C) to structure aware reasoning under absence (M). Current object detectors perform well when the task aligns with their underlying assumptions (C), but degrade significantly when those assumptions are not met (M). The combined setting (A) further reveals that these limitations are not isolated, but interact in realistic scenarios.
\section{Conclusion}

We introduced PCB-MC, a dataset for missing component detection with footprint level annotations and unified evaluation settings.

Our results show that detecting missing components is significantly more challenging than standard object detection, with high false negative rates across modern detectors and complete failure of anomaly methods.

Further research will explore structure aware modeling, including spatial expectations, layout priors, or relational reasoning. We expect PCB-MC to serve as a benchmark to stimulate research in this direction in industrial inspection.

\bibliographystyle{ieeetr}
\bibliography{main}

@String(CVPR= {IEEE Conf. Comput. Vis. Pattern Recog.})

@String(ICPR = {Int. Conf. Pattern Recog.})

@String(ICIP = {IEEE Int. Conf. Image Process.})

@String(CVPR  = {CVPR})

@String(ICPR  = {ICPR})

@String(ICIP  = {ICIP})

@article{Ling2023,
	title = {Printed {Circuit} {Board} {Defect} {Detection} {Methods} {Based} on {Image} {Processing}, {Machine} {Learning} and {Deep} {Learning}: {A} {Survey}},
	volume = {11},
	copyright = {https://creativecommons.org/licenses/by/4.0/legalcode},
	issn = {2169-3536},
	shorttitle = {Printed {Circuit} {Board} {Defect} {Detection} {Methods} {Based} on {Image} {Processing}, {Machine} {Learning} and {Deep} {Learning}},
	url = {https://ieeexplore.ieee.org/document/10044670/},
	doi = {10.1109/ACCESS.2023.3245093},
	urldate = {2025-11-03},
	journal = {IEEE Access},
	author = {Ling, Qin and Isa, Nor Ashidi Mat},
	year = {2023},
	pages = {15921--15944},
}

@article{Chhetri2023,
	title = {Detection of {Missing} {Component} in {PCB} {Using} {YOLO}},
	volume = {1},
	issn = {3021-940X},
	url = {https://www.nepjol.info/index.php/injet/article/view/60902},
	doi = {10.3126/injet.v1i1.60902},
	number = {1},
	urldate = {2025-11-03},
	journal = {International Journal on Engineering Technology},
	author = {Chhetri, Shivaji Pandit and Bhat, Santosh and Timalsina, Pradeep and Magar, Bipin Thapa},
	month = dec,
	year = {2023},
	pages = {62--71},
}

@article{Chavan2016QualityCO,
	title = {Quality {Control} of {PCB} using {Image} {Processing}},
	volume = {141},
	issn = {09758887},
	url = {http://www.ijcaonline.org/archives/volume141/number5/chavan-2016-ijca-909623.pdf},
	doi = {10.5120/ijca2016909623},
	number = {5},
	urldate = {2025-11-03},
	journal = {International Journal of Computer Applications},
	author = {R., Rasika and A., Swati and D., Gautami and B., Mayuri and S.Vaidya, Archana},
	month = may,
	year = {2016},
	pages = {28--32},
}

@INPROCEEDINGS{Demir1994,
author ={Demir, D. and Birecik, S. and Kurugollu, F. and Sezgin, M. and Bucak, I.O. and Sankur, B. and Anarim, E.},
booktitle ={20th Annual Conference of IEEE Industrial Electronics}, 
title ={Quality inspection in PCBs and SMDs using computer vision techniques}, 
year ={1994},
volume ={2},
number ={},
pages ={857-861 vol.2},
doi = {10.1109/IECON.1994.397899}}

@article{FasterRCNN,
  author       = {Shaoqing Ren and
                  Kaiming He and
                  Ross B. Girshick and
                  Jian Sun},
  title        = {Faster {R-CNN:} Towards Real-Time Object Detection with Region Proposal
                  Networks},
  journal      = {CoRR},
  volume       = {abs/1506.01497},
  year         = {2015},
  url          = {http://arxiv.org/abs/1506.01497},
  eprinttype   = {arXiv},
  eprint       = {1506.01497},
  bibsource    = {dblp computer science bibliography, https://dblp.org}
}

@article{MaskRCNN,
  author       = {Kaiming He and
                  Georgia Gkioxari and
                  Piotr Doll{\'{a}}r and
                  Ross B. Girshick},
  title        = {Mask {R-CNN}},
  journal      = {CoRR},
  volume       = {abs/1703.06870},
  year         = {2017},
  url          = {http://arxiv.org/abs/1703.06870},
  eprinttype   = {arXiv},
  eprint       = {1703.06870},
  bibsource    = {dblp computer science bibliography, https://dblp.org}
}

@inproceedings{Suksukont2025,
	
	title = {A {Deep} {Learning}-{Based} {System} for {Detecting} {Defects} in {Printed} {Circuit} {Boards}},
	copyright = {https://doi.org/10.15223/policy-029},
	isbn = {979-8-3315-4395-2},
	url = {https://ieeexplore.ieee.org/document/10987887/},
	doi = {10.1109/iEECON64081.2025.10987887},
	urldate = {2025-11-03},
	booktitle = {13th {International} {Electrical} {Engineering} {Congress} ({iEECON})},
	publisher = {IEEE},
	author = {Suksukont, Aekkarat and Onshaunjit, Jakkrit and Srinonchat, Jakkree},
	month = mar,
	year = {2025},
	pages = {1--6},
}

@inproceedings{Savu2025,

	title = {Reference-{Based} {Detection} and {Classification} of {Printed} {Circuit} {Boards} {Defects} {Using} {Deep} {Learning} and {Image} {Processing} {Techniques}},
	copyright = {https://doi.org/10.15223/policy-029},
	isbn = {979-8-3315-3352-6},
	url = {https://ieeexplore.ieee.org/document/11095497/},
	doi = {10.1109/ECAI65401.2025.11095497},
	urldate = {2025-11-03},
	booktitle = {17th {International} {Conference} on {Electronics}, {Computers} and {Artificial} {Intelligence} ({ECAI})},
	publisher = {IEEE},
	author = {Savu, Andreea-Daniela and Bizon, Nicu and Dragusin, Sebastian-Alexandru},
	month = jun,
	year = {2025},
	pages = {1--10},
}

@article{Makwana2023,
	title = {Pcbsegclassnet - a {Light}-{Weight} {Network} for {Segmentation} and {Classification} of {Pcb} {Component}},
	issn = {1556-5068},
	url = {https://www.ssrn.com/abstract=4241188},
	doi = {10.2139/ssrn.4241188},
	language = {en},
	urldate = {2025-11-03},
	journal = {SSRN Electronic Journal},
	author = {Makwana, Dhruv and R, Sai Chandra Teja and Mittal, Sparsh},
	year = {2022},
}

@article{Lu2020,
  title={FICS-PCB: A Multi-Modal Image Dataset for Automated Printed Circuit Board Visual Inspection},
  author={Hangwei Lu and Dhwani Mehta and Olivia P. Paradis and Navid Asadizanjani and Mark Mohammad Tehranipoor and D. Woodard},
  journal={IACR Cryptol. ePrint Arch.},
  year={2020},
  volume={2020},
  pages={366},
  url={https://api.semanticscholar.org/CorpusID:215796830}
}

@inproceedings{Mahalingam2015,
	
	title = {{PCB}-{METAL}: {A} {PCB} {Image} {Dataset} for {Advanced} {Computer} {Vision} {Machine} {Learning} {Component} {Analysis}},
	isbn = {978-4-901122-18-4},
	shorttitle = {{PCB}-{METAL}},
	url = {https://ieeexplore.ieee.org/document/8757928/},
	doi = {10.23919/MVA.2019.8757928},
	urldate = {2025-11-03},
	booktitle = {16th {International} {Conference} on {Machine} {Vision} {Applications} ({MVA})},
	publisher = {IEEE},
	author = {Mahalingam, Gayathri and Gay, Kevin Marshall and Ricanek, Karl},
	month = may,
	year = {2019},
	pages = {1--5},
}

@inproceedings{Pramerdorfer2015,
	
	title = {A dataset for computer-vision-based {PCB} analysis},
	isbn = {978-4-901122-14-6},
	url = {http://ieeexplore.ieee.org/document/7153209/},
	doi = {10.1109/MVA.2015.7153209},
	urldate = {2025-11-03},
	booktitle = {14th {IAPR} {International} {Conference} on {Machine} {Vision} {Applications} ({MVA})},
	publisher = {IEEE},
	author = {Pramerdorfer, Christopher and Kampel, Martin},
	month = may,
	year = {2015},
	pages = {378--381},
}

@article{LV2024,
	title = {A dataset for deep learning based detection of printed circuit board surface defect},
	volume = {11},
	issn = {2052-4463},
	url = {https://www.nature.com/articles/s41597-024-03656-8},
	doi = {10.1038/s41597-024-03656-8},
	language = {en},
	number = {1},
	urldate = {2025-11-03},
	journal = {Scientific Data},
	author = {Lv, Shengping and Ouyang, Bin and Deng, Zhihua and Liang, Tairan and Jiang, Shixin and Zhang, Kaibin and Chen, Jianyu and Li, Zhuohui},
	month = jul,
	year = {2024},
	pages = {811},
}

@article{Li2019,
	title = {{TDD}‐net: a tiny defect detection network for printed circuit boards},
	volume = {4},
	issn = {2468-6557, 2468-2322},
	shorttitle = {{TDD}‐net},
	url = {https://ietresearch.onlinelibrary.wiley.com/doi/10.1049/trit.2019.0019},
	doi = {10.1049/trit.2019.0019},
	language = {en},
	number = {2},
	urldate = {2025-11-03},
	journal = {CAAI Transactions on Intelligence Technology},
	author = {Ding, Runwei and Dai, Linhui and Li, Guangpeng and Liu, Hong},
	month = jun,
	year = {2019},
	pages = {110--116},
}

@misc{Tang2018,
	title = {Online {PCB} {Defect} {Detector} {On} {A} {New} {PCB} {Defect} {Dataset}},
	copyright = {arXiv.org perpetual, non-exclusive license},
	url = {https://arxiv.org/abs/1902.06197},
	doi = {10.48550/ARXIV.1902.06197},
	urldate = {2025-11-03},
	publisher = {arXiv},
	author = {Tang, Sanli and He, Fan and Huang, Xiaolin and Yang, Jie},
	year = {2019},
	note = {Version Number: 1},
}

@misc{RF100,
Author = {Floriana Ciaglia and Francesco Saverio Zuppichini and Paul Guerrie and Mark McQuade and Jacob Solawetz},
Title = {Roboflow 100: A Rich, Multi-Domain Object Detection Benchmark},
Year = {2022},
Eprint = {arXiv:2211.13523},
}

@misc{lin2018focallossdenseobject,
      title={Focal Loss for Dense Object Detection}, 
      author={Tsung-Yi Lin and Priya Goyal and Ross Girshick and Kaiming He and Piotr Dollár},
      year={2018},
      eprint={1708.02002},
      archivePrefix={arXiv},
      primaryClass={cs.CV},
      url={https://arxiv.org/abs/1708.02002}, 
}

@misc{zhao2024detrsbeatyolosrealtime,
      title={DETRs Beat YOLOs on Real-time Object Detection}, 
      author={Yian Zhao and Wenyu Lv and Shangliang Xu and Jinman Wei and Guanzhong Wang and Qingqing Dang and Yi Liu and Jie Chen},
      year={2024},
      eprint={2304.08069},
      archivePrefix={arXiv},
      primaryClass={cs.CV},
      url={https://arxiv.org/abs/2304.08069}, 
}

@article{Tang2025LRTDETRAE,
  title={LRT-DETR: An Enhanced Transformer-Based Framework for PCB Surface Defect Detection},
  author={Hongyu Tang and Dengli Bu and Guanhao Mo and Panlin Lu and Zhenhai Li and Long Huang},
  journal={IEEE 6th International Seminar on Artificial Intelligence, Networking and Information Technology (AINIT)},
  year={2025},
  pages={1-10},
}

@INPROCEEDINGS{SAHI,
  author={Akyon, Fatih Cagatay and Onur Altinuc, Sinan and Temizel, Alptekin},
  booktitle={2022 IEEE International Conference on Image Processing (ICIP)}, 
  title={Slicing Aided Hyper Inference and Fine-Tuning for Small Object Detection}, 
  year={2022},
  volume={},
  number={},
  pages={966-970},
  doi={10.1109/ICIP46576.2022.9897990}}

@misc{peng2024dfineredefineregressiontask,
      title={D-FINE: Redefine Regression Task in DETRs as Fine-grained Distribution Refinement}, 
      author={Yansong Peng and Hebei Li and Peixi Wu and Yueyi Zhang and Xiaoyan Sun and Feng Wu},
      year={2024},
      eprint={2410.13842},
      archivePrefix={arXiv},
      primaryClass={cs.CV},
      url={https://arxiv.org/abs/2410.13842}, 
}

@software{yolov8_ultralytics,
  author = {Glenn Jocher and Ayush Chaurasia and Jing Qiu},
  title = {Ultralytics YOLOv8},
  version = {8.0.0},
  year = {2023},
  url = {https://github.com/ultralytics/ultralytics},
  orcid = {0000-0001-5950-6979, 0000-0002-7603-6750, 0000-0003-3783-7069},
  license = {AGPL-3.0}
}

@software{yolo11_ultralytics,
  author = {Glenn Jocher and Jing Qiu},
  title = {Ultralytics YOLO11},
  version = {11.0.0},
  year = {2024},
  url = {https://github.com/ultralytics/ultralytics},
  orcid = {0000-0001-5950-6979, 0000-0003-3783-7069},
  license = {AGPL-3.0}
}

@software{yolo26_ultralytics,
  author = {Glenn Jocher and Jing Qiu},
  title = {Ultralytics YOLO26},
  version = {26.0.0},
  year = {2026},
  url = {https://github.com/ultralytics/ultralytics},
  orcid = {0000-0001-5950-6979, 0000-0003-3783-7069},
  license = {AGPL-3.0}
}

@article{xia_global_2023,
	title = {Global contextual attention augmented {YOLO} with {ConvMixer} prediction heads for {PCB} surface defect detection},
	volume = {13},
	issn = {2045-2322},
	url = {https://www.nature.com/articles/s41598-023-36854-2},
	doi = {10.1038/s41598-023-36854-2},
	language = {en},
	number = {1},
	urldate = {2026-02-19},
	journal = {Scientific Reports},
	author = {Xia, Kewen and Lv, Zhongliang and Liu, Kang and Lu, Zhenyu and Zhou, Chuande and Zhu, Hong and Chen, Xuanlin},
	month = jun,
	year = {2023},
	pages = {9805},
}

@article{calabrese_application_2025,
	title = {Application of {Mask} {R}-{CNN} and {YOLOv8} algorithms for defect detection in printed circuit board manufacturing},
	volume = {7},
	issn = {3004-9261},
	url = {https://link.springer.com/10.1007/s42452-025-06641-x},
	doi = {10.1007/s42452-025-06641-x},
	language = {en},
	number = {4},
	urldate = {2026-02-19},
	journal = {Discover Applied Sciences},
	author = {Calabrese, Maurizio and Agnusdei, Leonardo and Fontana, Gianmauro and Papadia, Gabriele and Del Prete, Antonio},
	month = mar,
	year = {2025},
	pages = {257},
}

@inproceedings{roth2022patchcore,
  title={Towards Total Recall in Industrial Anomaly Detection},
  author={Roth, Karsten and Pemula, Latha and Zepeda, Joaqu{\'i}n and Scholkopf, Bernhard and Brox, Thomas and Gehler, Peter},
  booktitle={CVPR},
  year={2022}
}

@inproceedings{defard2021padim,
  title={PaDiM: a Patch Distribution Modeling Framework for Anomaly Detection and Localization},
  author={Defard, Thomas and Setkov, Alexey and Loesch, Angelique and Audigier, Romain},
  booktitle={ICPR},
  year={2021}
}

@misc{zavrtanik2021draemdiscriminativelytrained,
      title={DRAEM -- A discriminatively trained reconstruction embedding for surface anomaly detection}, 
      author={Vitjan Zavrtanik and Matej Kristan and Danijel Skočaj},
      year={2021},
      eprint={2108.07610},
      archivePrefix={arXiv},
      primaryClass={cs.CV},
      url={https://arxiv.org/abs/2108.07610}, 
}

@inproceedings{deng2022reverse, 
title = {Anomaly Detection via Reverse Distillation from One-Class Embedding}, 
author = {Deng, Hanqiu and Li, Xingyu}, 
booktitle = {Proc. IEEE/CVF Conf. Comput. Vis. Pattern Recognit. (CVPR)}, 
pages = {9727--9736}, 
year = {2022} }

@misc{lin2015microsoftcococommonobjects,
      title={Microsoft COCO: Common Objects in Context}, 
      author={Tsung-Yi Lin and Michael Maire and Serge Belongie and Lubomir Bourdev and Ross Girshick and James Hays and Pietro Perona and Deva Ramanan and C. Lawrence Zitnick and Piotr Dollár},
      year={2015},
      eprint={1405.0312},
      archivePrefix={arXiv},
      primaryClass={cs.CV},
      url={https://arxiv.org/abs/1405.0312}, 
}

@misc{loshchilov2019decoupledweightdecayregularization,
      title={Decoupled Weight Decay Regularization}, 
      author={Ilya Loshchilov and Frank Hutter},
      year={2019},
      eprint={1711.05101},
      archivePrefix={arXiv},
      primaryClass={cs.LG},
      url={https://arxiv.org/abs/1711.05101}, 
}

@article{Kiobya2024ACASEM,
  title={Attentive context and semantic enhancement mechanism for printed circuit board defect detection with two-stage and multi-stage object detectors},
  author={Kiobya, Twahir and Zhou, Junfeng and Maiseli, B. and Khan, Maqbool},
  journal={Scientific Reports},
  year={2024},
  doi={10.1038/s41598-024-69207-8},
  url={https://www.nature.com/articles/s41598-024-69207-8.pdf}
}

@article{Hu2020PCBFasterRCNN,
  title={Detection of PCB Surface Defects With Improved Faster-RCNN and Feature Pyramid Network},
  author={Hu, Bing and Wang, Jianhui},
  journal={IEEE Access},
  year={2020},
  doi={10.1109/ACCESS.2020.3001349},
  url={https://ieeexplore.ieee.org/ielx7/6287639/8948470/09113299.pdf}
}

@article{Luo2021DecoupledTwoStage,
  title={FPCB Surface Defect Detection: A Decoupled Two-Stage Object Detection Framework},
  author={Luo, Jiaxiang and Yang, Zhiyu and Li, Shipeng and Wu, Yilin},
  journal={IEEE Transactions on Instrumentation and Measurement},
  year={2021},
  doi={10.1109/TIM.2021.3092510},
  url={https://www.semanticscholar.org/paper/e5448da0aa628e009b4025251a4e50922afe6ad4}
}

@article{DBLP:journals/corr/HeZRS15,
  author       = {Kaiming He and
                  Xiangyu Zhang and
                  Shaoqing Ren and
                  Jian Sun},
  title        = {Deep Residual Learning for Image Recognition},
  journal      = {CoRR},
  volume       = {abs/1512.03385},
  year         = {2015},
  url          = {http://arxiv.org/abs/1512.03385},
  eprinttype   = {arXiv},
  eprint       = {1512.03385},
  bibsource    = {dblp computer science bibliography, https://dblp.org}
}

@article{liu2024deep, 
title = {Deep Industrial Image Anomaly Detection: A Survey}, 
author = {Liu, Jiaqi and Xie, Guoyang and Wang, Jinbao and Li, Shangnian and Wang, Chengjie and Zheng, Feng and Jin, Yaochu},
journal = {Mach. Intell. Res.},
volume = {21}, 
number = {1},
pages = {104--135},
year = {2024} }

@InProceedings{bergmann2019mvtec, 
author = {Bergmann, Paul and L\"owe, Sindy and Fauser, Michael and Sattlegger, David and Steger, Carsten}, 
title = {MVTec AD -- A Comprehensive Real-World Dataset for Unsupervised Anomaly Detection}, 
booktitle = {Proceedings of the IEEE/CVF Conference on Computer Vision and Pattern Recognition (CVPR)},
month = {June},
year = {2019} }

@inproceedings{akcay2022anomalib, 
title={Anomalib: A Deep Learning Library for Anomaly Detection}, 
author={Akcay, Samet and Ameln, Dick and Vaidya, Ashwin and Lakshmanan, Barath and Ahber, Nilesh and Genc, Utku}, booktitle={IEEE International Conference on Image Processing (ICIP)}, 
year={2022} }

\end{document}